\documentclass[conference, letterpaper]{IEEEtran}
\usepackage{subcaption}
\usepackage[colorlinks=true,
            linkcolor=red,
            urlcolor=blue,
            breaklinks=true,
            citecolor=blue]{hyperref}
\usepackage[ruled,vlined]{algorithm2e}
\usepackage{mathtools}
\usepackage{xcolor}
\usepackage{sidecap}
\usepackage{graphicx,float}
\usepackage{balance}

\usepackage{fancyhdr}

\renewcommand{\thispagestyle}[2]{}

\fancypagestyle{plain}{
        \fancyhead{}
        \fancyhead[C]{first page center header}
        \fancyfoot{}
        \fancyfoot[C]{first page center footer}
}
\begin{document}

\title{Weighted Unsupervised Learning for 3D Object Detection}

\author{\IEEEauthorblockN{Kamran Kowsari}
\IEEEauthorblockA{Department of Computer Science,\\The George Washington University, Washington DC, \\
 \\}
\and
\IEEEauthorblockN{Manal H. Alassaf}
\IEEEauthorblockA{Department of Computer Science,\\The George Washington University, Washington DC\\ Department of Computer Science,\\ Taif University, Taif, Saudi Arabia \\
\\}

 }

%Telephone: (800) 555--1212, Fax: (888) 555--1212}


\maketitle

\begin{abstract}
%\boldmath
This paper introduces a novel weighted unsupervised learning for object detection using an RGB-D camera. This technique is feasible for detecting the moving objects in the noisy environments that are captured by an RGB-D camera. The main contribution of this  paper is a real-time algorithm for detecting each object using weighted clustering as a separate cluster. In a preprocessing step, the algorithm calculates the pose 3D position X, Y, Z and RGB color of each data point and then it calculates each data point's normal vector using the point's neighbor. After preprocessing, our algorithm calculates k-weights for each data point; each weight indicates membership. Resulting in clustered objects of the scene.
\end{abstract}



% \fi
\IEEEpeerreviewmaketitle

\begin{figure*}
  \includegraphics[width=2.1\columnwidth]{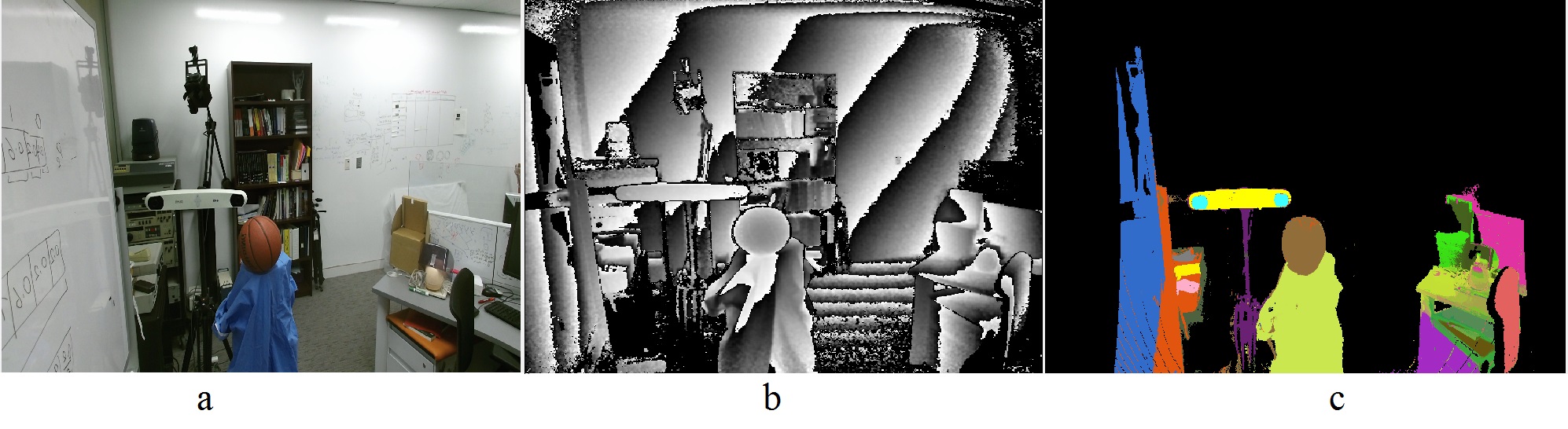}
\caption{ $a$) Kinect color frame (RGB) with resolution of 1920 X 1080; $b$) Kinect depth frame with resolution of 512 X 424;\\$c$) Proposed method object detection using k= 15 clusters, and after 15 iterations.}
\label{fig:1} 
\end{figure*}
\section{Introduction}
\label{intro}
 Object detection for unlabeled and unsegmented data points~\cite{RefWorks:74} is widely studied. In general, visualization and machine learning are the main issues, which are reviewed in existing studies. Machine learning is classified into two categories; supervised and unsupervised learning. In the first case, many researches have addressed the object detection with supervised methods~\cite{RefWorks:115,RefWorks:83,RefWorks:113,RefWorks:107,RefWorks:118,RefWorks:119,RefWorks:127}. Kevin Lai's and many other researchers work on the weighted supervised learning for object detection~\cite{RefWorks:114,RefWorks:15,RefWorks:88,RefWorks:89} using a hierarchical, multi views, and sparse distance learning. That method can be useful for known objects intended to be detected. Therefore, his algorithm is used for object detection of a specific item that is stored in a database in a preprocessing step such as an apple, egg, etc. Therefore, when using supervised learning for object detection, researchers focus on the accuracy of object detection and detecting known objects. On the other hand, time is very critical in real-time object detection techniques.

The second category addresses the object detection problem with unsupervised learning method~\cite{RefWorks:115,RefWorks:100,RefWorks:84,RefWorks:86,RefWorks:106,RefWorks:72,RefWorks:122,RefWorks:81,RefWorks:105,RefWorks:111} using techniques of unsupervised learning such as K-means and spectral clustering. Most of these techniques are not sufficient for real-time applications since they do not address time complexity and memory consumption. With regard to accuracy of object detection using an RGB-D camera, an efficient method for RGB-D camera is weighted clustering since capturing by this kind of camera has more noise introduced by moving objects; thus, labels need to be updated in a few iterations that span less than a second. If we want to compare the clustering between computer graphics and other domains; data points are not changing during running time in most fields such as data mining, but data points in real-time object detection in the field of computer graphics, machine vision, and robotics are continuously changing; frame by frame and second by second. As regards to Liefeng Bo~\cite{RefWorks:84} who uses a dictionary for his work, this technique is very efficient for synchronized video, and also this method is fast enough for video processing with around 2 frames per second (FPS). But, in real-time applications, it needs to be faster than 2 FPS. Weighted Unsupervised Learning such as weighted k-means~\cite{RefWorks:85} or many other methods~\cite{RefWorks:50,RefWorks:cluster,RefWorks:10,RefWorks:1010,RefWorks:1011,RefWorks:1012} are implemented in different domains. Fuzzy object detection and weighted clustering is addressed and used for moving objects~\cite{RefWorks:86,RefWorks:130,RefWorks:128}. In 2010, Maddalena et al.~\cite{RefWorks:86}  had worked on fuzzy logics and learning. That work used only pixel-by-pixel as their features, which can be very efficient for image processing. Thus, those methods never use other sensors such as depth information as a specific feature for 3D, whereas Maddalena's method has many limitations for indoor capturing for 3D object detection. Some other researches use Vicon system as object detection and object controlling for real-time applications, but this device cannot capture color and surface of the objects~\cite{RefWorks:109,RefWorks:110,RefWorks:125,RefWorks:1250,RefWorks:1251} ; thus, Vicon cannot be implemented as a colorful application; however, it can be an efficient method for rigid object detection that could calculate object position as the only feature. Therefore, an unsupervised and RGB-D camera method that addresses the accuracy, time-complexity, and memory consumption and colorful surface capturing simultaneously is unprecedented.

Visualization is an important step in object detection for evaluating the results. Graphical Processing Unit (GPU) based application is a powerful method in the computer graphic domain, but hence not applicable in mobility applications. In general, GPU-based application needs powerful hardware; therefore, for mobility application, researchers focus more on Central Processing Unit (CPU) based applications.

This paper is an extension of the authors' prior work in~\cite{RefWorks:100} where we used RGB-D camera and studied the Boolean version of clustering. The color space from RGB was improved to Hunter-Lab color space~\cite{RefWorks:90,RefWorks:91}  where the Hunter-Lab color space gave smoother result of clustering in real-time application. In this paper, we use the RGB color space. We improved the clustering part by adding k weights to each data point. This improvement affects frame rate, accuracy, and memory consumption in the real-time application. We propose a real-time object detection algorithm using k weighted clusters with memory consumption that is useful for mobility application. The maximum memory needed for this algorithm is 650 MB for 50 clusters, less than one GB for 100 clusters, and we use multi-thread processing to improve frame rate.

In short, new contributions and unique features of the proposed method in this paper are as follows. 1. Weighted unsupervised learning is presented, which reduces noise for moving and small objects, has better time complexity, lower memory consumption, and higher accuracy than previous methods, 2. CPU-based implementation is offered to make our method capable of mobility usage, and 3. A segmentation technique is proposed to detect a user defined object.

%%%%%%%%%%%%%%%%%%%%%%%%%%%%%%%%%%%%%%%%%%%%%%%%%%%%%%%%%%%%%%%%%%%%%%%%%
%%%%%%%%%%%%%%%%%%%%%%%%%%%%%%%%%%%%%%%%%%%%%%%%%%%%%%%%%%%%%%%%%%%%%%%%%

\begin{figure*}[ht]
 \centering \includegraphics[width=2.0 \columnwidth]{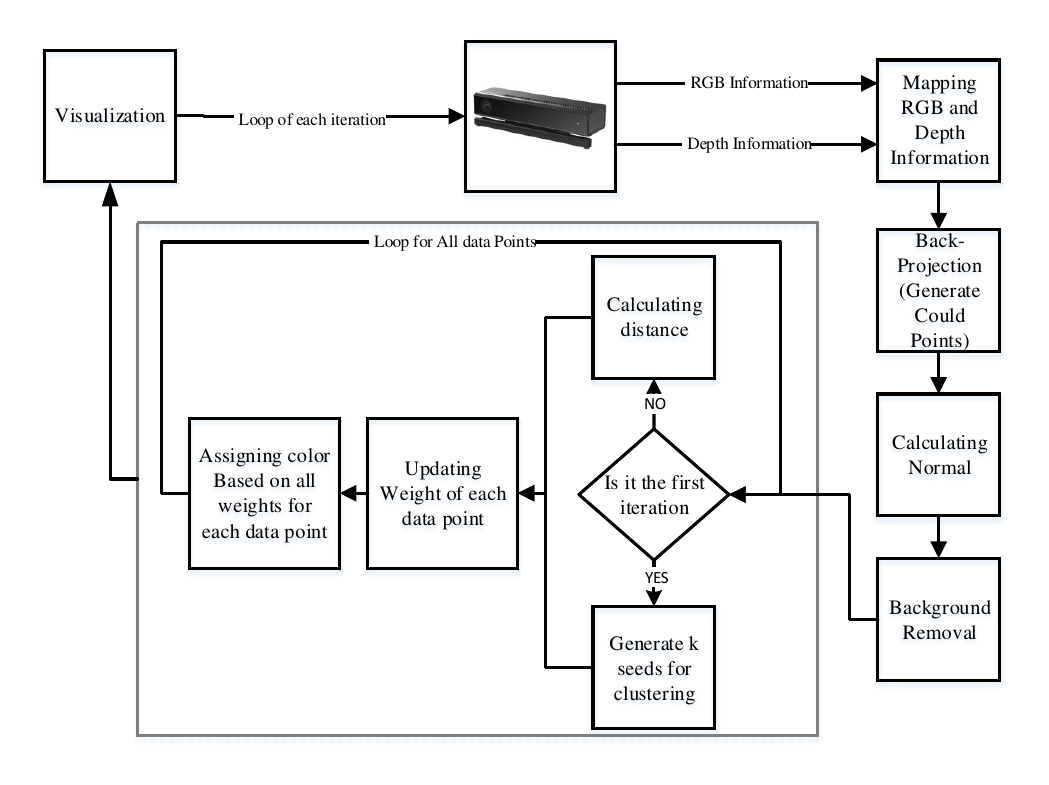}
\caption{Pipeline of 3D Object detection using RGB-D camera has two main parts: 1) Preprocessing including Mapping, Back- Projection, Normal Generating, Background removal and 2) Clustering including assigned initial weight, distance calculation, update weight and assign color, and finally visualization to illustrate the results.}
\label{fig:2} 
\end{figure*}

\section{Pipeline and Methods}\label{sec:Pipeline}
The pipeline of weighed unsupervised object detection algorithm presented in this paper is illustrated in Figure~\ref{fig:2} and composed of 8 steps as follows: 
\begin{enumerate}
\item Capturing RGB color and depth information; 
\item Mapping RGB and depth information; 
\item Applying  back-projection for generating cloud points in the 3D world coordinate system; 
\item Calculating data points normal vectors based on each point neighbors;
\item Segmenting and removing the background to limit the area where we want to detect the objects;
\item Calculating distance between each point and the k-centers of the clusters;
\item Updating the k-weights of each cloud point; and
\item Assigning color to each data point by using the point's k-weights. 
\end{enumerate}

This paper is organized as follows: preprocessing is presented in section~\ref{sec:pre}  followed by clustering step in section ~\ref{sec:clustering}. Section ~\ref{sec:Visualization} talks about visualization. Finally, numerical and experimental results are presented in section ~\ref{sec:Results}. Followed by discussion in section ~\ref{sec:Discussion} and finally the conclusion and future work is in section ~\ref{sec:Conclusion}. 
%%%%%%%%%%%%%%%%%%%%%%%%%%%%%%%%%%%%%%%%%%%%%%%%%%%%%%%%%%%%%%%%%%%%%%%%%

\subsection{Preprocessing}\label{sec:pre}
Preprocessing is used for generating the 3D cloud points from input data that is captured by an RGB-D camera and needed for clustering steps; thus, in preprocessing steps, the algorithm generates cloud points by pose-the 3D as XYZ, color as RGB, and normal as $n_x$, $n_y$, $n_z$. In this research paper, we use Kinect for Windows V2 as an RGB-D camera. In short, preprocessing steps are: get input, perform frames mapping, back projection, and normal calculation.
%%%%%%%%%%%%%%%%%%%%%%%%%%%%%%%%%%%%%%%%%%%%%%%%%%%%%%%%%%%%%%%%%%%%%%%%%
\subsubsection{Get Input}\label{sec:input}
The Kinect camera was designed as a hands-free game controller. It has two input sensors, which include an RGB camera with a resolution of 1920 X 1080 pixels, and a depth sensor with the resolution of 512 X 424 pixels. Field of View (FOV) is 84.1 X 53.8 for RGB color space and 70.6 X 60 degree for depth sensor information. The resulting average is about 22 X 20 pixels per degree for RGB and 7 X 7 pixels per degree of depth data ~\cite{RefWorks:92,RefWorks:93}. Kinect can capture depth information of objects displaced up to 4.5-5 meters from the camera, but we can limit it manually between a and b meters for indoor object detection~\cite{RefWorks:75,RefWorks:95}. 
%%%%%%%%%%%%%%%%%%%%%%%%%%%%%%%%%%%%%%%%%%%%%%%%%%%%%%%%%%%%%%%%%%%%%%%%%
\subsubsection{Frames Mapping}\label{sec:mapping}
The two inputs' frames of Kinect have dissimilar resolutions. The mapping  is needed  for matching color space and depth information in the same space. Our approach is like other studies~\cite{RefWorks:96,RefWorks:97} for generating colored cloud points. After mapping, we have aligned frames with information about the position x, y, z; and color R, G, and B for each data point. These data points will be ready for the next step, back-projection~\cite{RefWorks:96,RefWorks:97}. 
%%%%%%%%%%%%%%%%%%%%%%%%%%%%%%%%%%%%%%%%%%%%%%%%%%%%%%%%%%%%%%%%%%%%%%%%%
\subsubsection{Back Projection}\label{sec:back}
n this step we convert the cloud points into 3D world coordinate system. The Kinect input from mapping step includes x, y, and z parameters. We follow equations ~\ref{eq:10} through ~\ref{eq:14} that are demonstrated in the appendix to back-project the points into the world coordinate System. In summary, to generate an accurate 3D position of each data point in the color frame, the 2D position is back-projected using the depth data from the depth frame as z and the Kinect camera intrinsic parameters to obtain the correct 3D position of each cloud point in the real world coordinate system.
\newpage
%%%%%%%%%%%%%%%%%%%%%%%%%%%%%%%%%%%%%%%%%%%%%%%%%%%%%%%%%%%%%%%%%%%%%%%%%
\subsubsection{Normal Calculation}\label{sec:normal}
In this step, normal vector of each cloud point is calculated as new feature indicated by  $n_x$, $n_y$, $n_z$. The normal of each data point is calculated using its neighbors.
\begin{equation}
p_{i}=(x\; y\; z\; r\; g\; b\; n_{x}\; n_{y}\; n_{z})^{T}.
\label{eq:1} 
\end{equation}

%%%%%%%%%%%%%%%%%%%%%%%%%%%%%%%%%%%%%%%%%%%%%%%%%%%%%%%%%%%%%%%%%%%%%%%%%

\subsection{Clustering step}\label{sec:clustering}
The aim of this paper is object detection using weighted unsupervised learning. In our approach, we use k-means clustering algorithm with k-weights for each data point, where k is the number of clusters that is defined by use. The clustering step is divided into two main parts; initial seeding, and updating the weights. First part in the first iteration is initialization of k-seeds for the k-means algorithm, and initialization of k-weights for each data point. We use k-means++~\cite{RefWorks:98,RefWorks:1245} algorithm to obtain the initial seeds of the k-means clustering, while the k-weights for each data point are initialized to zero. The second part is updating the k-weights~\cite{RefWorks:85}. Weights are going to be updated at each of the following iterations. At each iteration, each data point will have k values which indicate a membership for one of the k clusters. At the end of each iteration, each point will belong to the cluster that has the highest weight. The algorithm details are given in equation~\ref{al:1}, where $C_i$ is the color of each cluster that is assigned at the beginning as unique color for all clusters. The $\mu_{i}$ is k weights of each point, and will continually be changing by each iteration and getting new values. Clustering step includes the labeling, updating the k weights for each data point, and finally visualization is utilized to illustrate the results.
%%%%%%%%%%%%%%%%%%%%%%%%%%%%%%%%%%%%%%%%%%%%%%%%%%%%%%%%%%%%%%%%%%%%%%%%%

%%%

\begin{figure*}[!ht]
  \includegraphics[width=2.1\columnwidth]{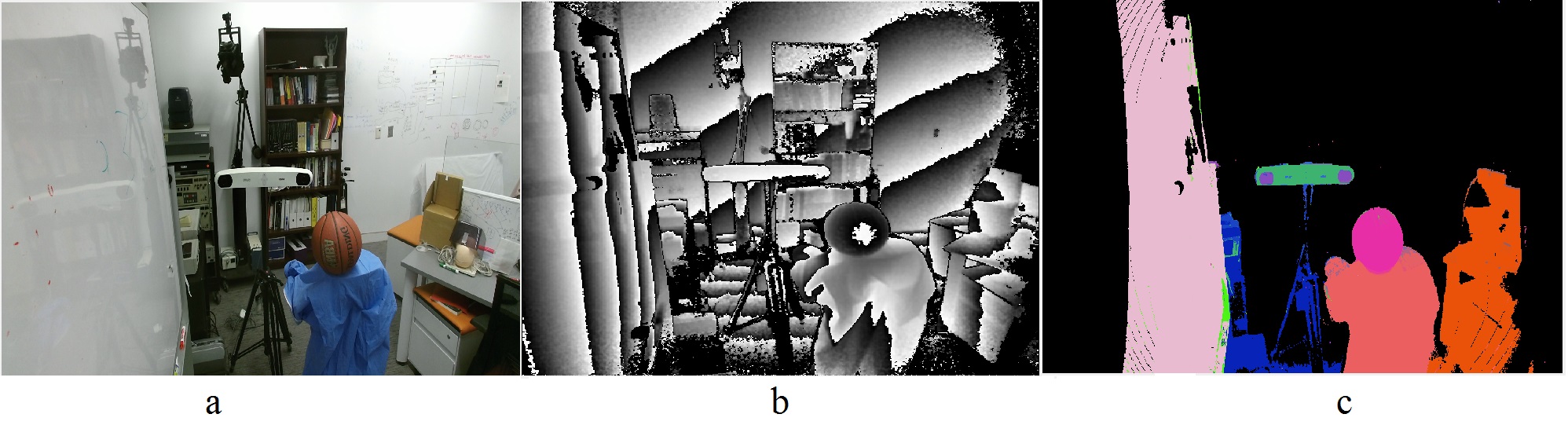}
\caption{ $a$) Kinect color frame (RGB) with resolution of 1920 X 1080; $b$) Kinect depth frame with resolution of 512 X 424;\\$c$) Proposed method object detection using k= 7 clusters, and after 10 iterations. Memory consumption is 320 MB and frame rate is 8.1 $\pm 0.2$  FPS.\\}
\label{fig:3} 
\end{figure*}

\begin{figure*}[!ht]
  \includegraphics[width=2.1\columnwidth]{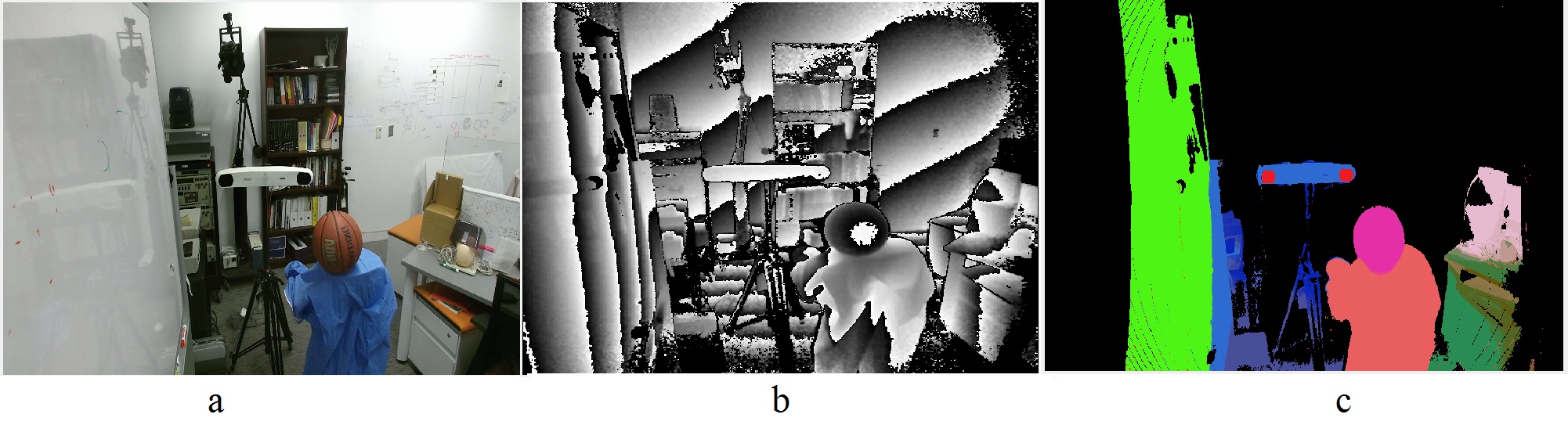}
\caption{ $a$) Kinect color frame (RGB) with resolution of 1920 X 1080; $b$) Kinect depth frame with resolution of 512 X 424;\\$c$) Proposed method object detection using k=10 clusters, and after10 iterations. Memory consumption is 340 MB and frame rate is  6.5$\pm 0.2$  FPS.\\}
\label{fig:4} 
\end{figure*}

\begin{figure*}[!ht]
  \includegraphics[width=2.1\columnwidth]{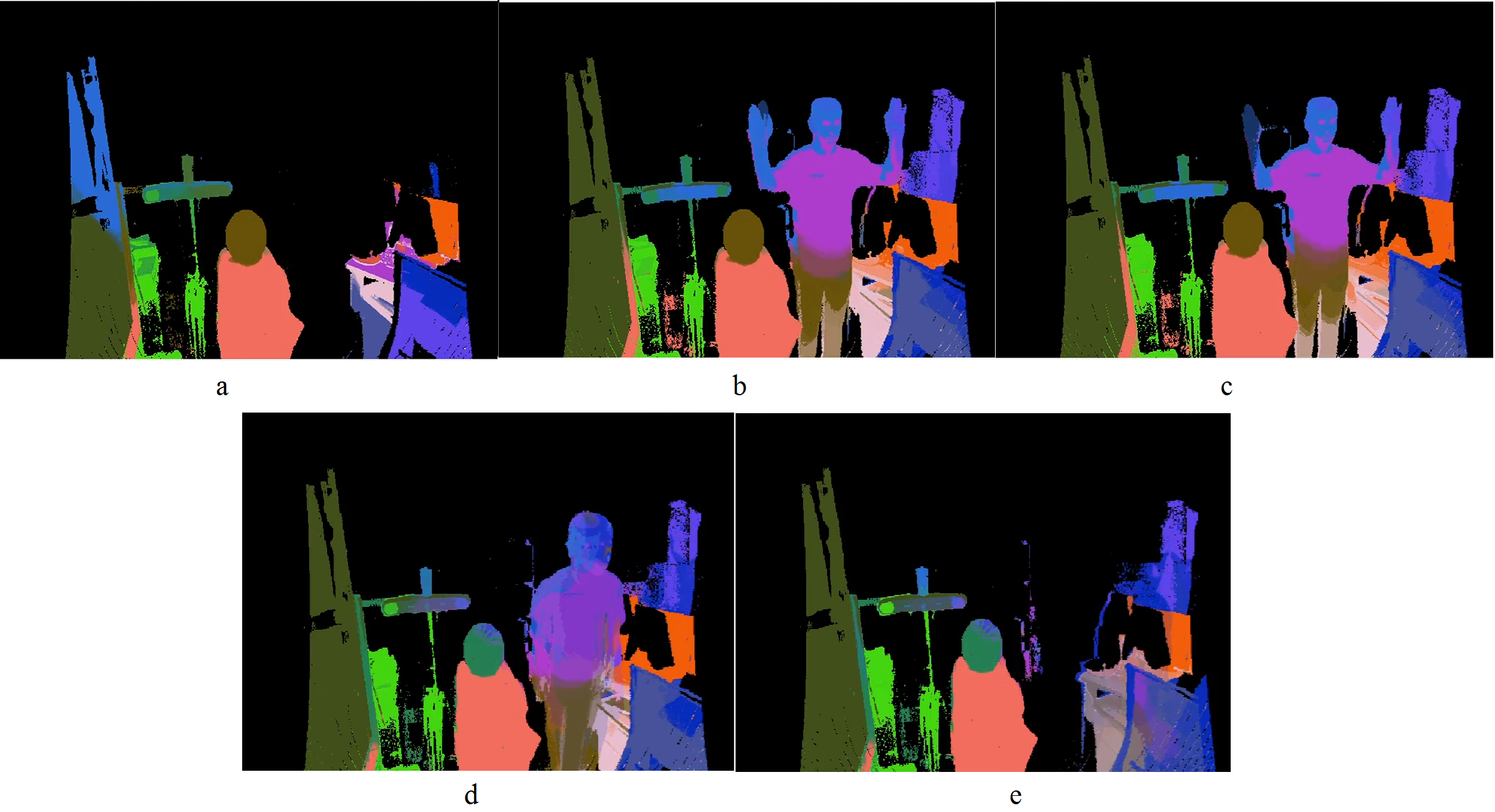}
\caption{ $a$) Shows clustered output after 10 iterations and using k=13 clusters; $b$) Shows a moving object entering the scene.  After several iterations the algorithm succeeded in distinguishing the moving object as new object; $c$) Shows moving hand, where the algorithm successfully detected the moving object; $d$) Shows a moving object going out of the scene and how the weight of that object was reduced only after one iteration, and finally; $e$) Shows clustered output after several iterations as detected objects. }
\label{fig:5} 
\end{figure*}

\begin{figure*}\center 
\includegraphics[width=1.15\columnwidth]{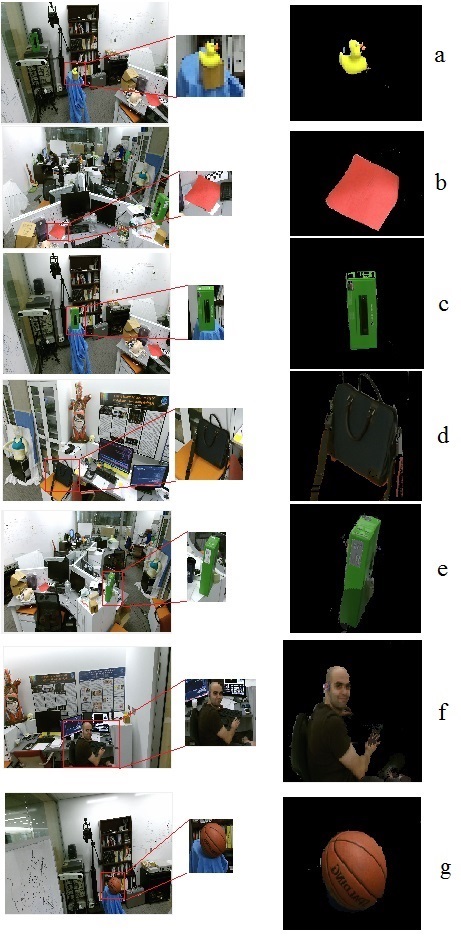}
\caption{  Results of segmenting scene objects using proposed algorithm; $a$) Segmentation of small duck; $ b$) Segmentation and detection of piece of red paper; $c$) Object detection of a box; $d$) Shows handy bag; $e$)  Segmentation of box, the border of the box has lower weight and it will be completed after several iteration; $f$) Representation of moving object, segmentation of a person; $g$) Segmentation of basketball. }

\label{fig:6} 
\end{figure*}

\begin{figure*}\center [!ht]
\includegraphics[width=2.0\columnwidth]{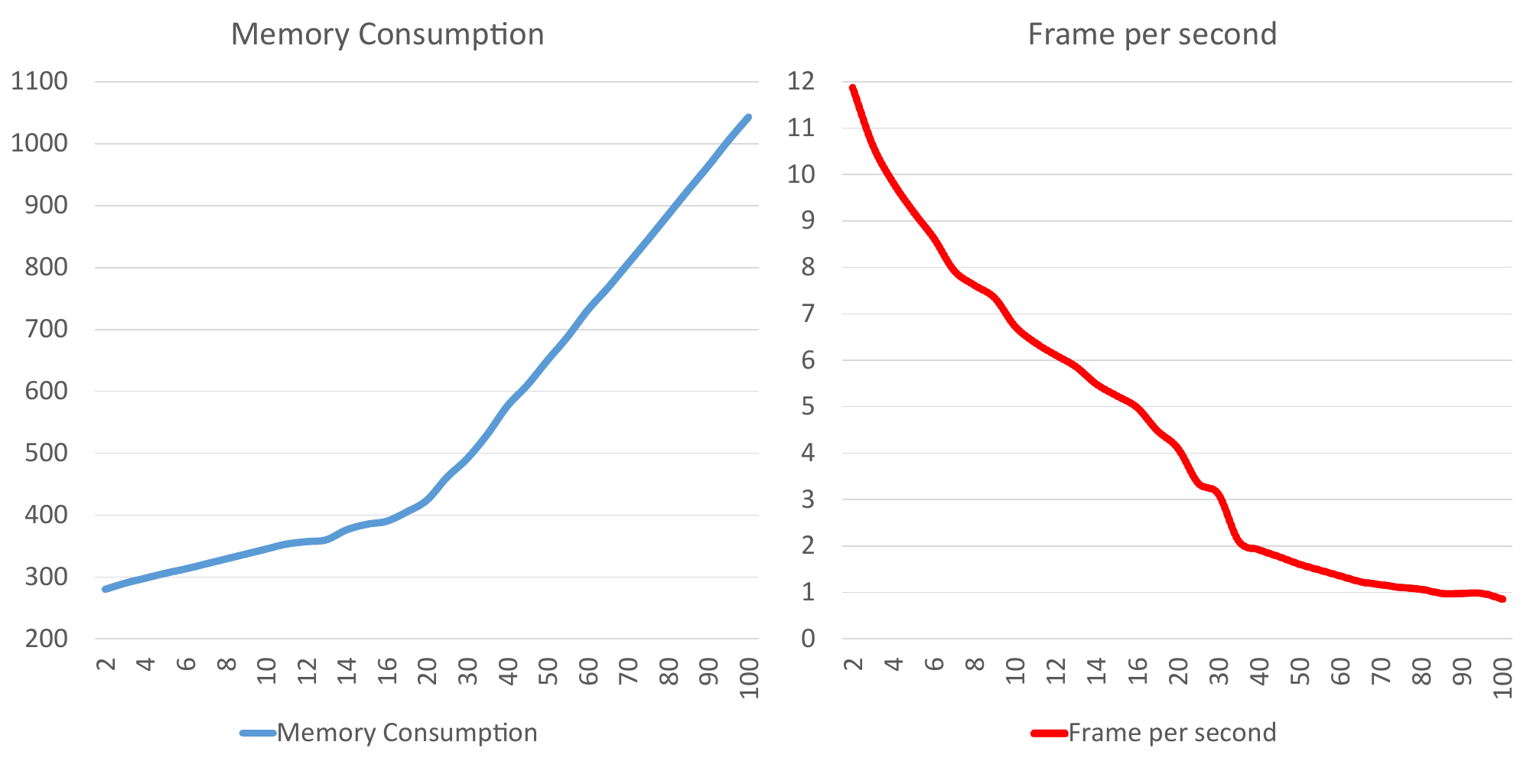}
\caption{ Left: Chart indicates the memory consumption for different number of clusters where x-axis represents the number of clusters, k, value and y-axis show the memory consumption in MB. The larger the k value the more memory needed to process all scene points. Right: The frame rate is given in the right chart where x-axis is number of clusters, k, and y-axis is frame per second. The larger the k value the less frames per second evaluated.}
\label{fig:7} 
\end{figure*}

%%%%%%%%%%%%%%%%%%%%%%%%%%%%%%%%%%%%%%%%%%%%%%%%%%%%%%%%%%%%%%%%%%%%%%%%%

\subsubsection{Labeling}\label{sec:Labeling}
In the adopted weighted unsupervised learning, each data point has a k-values. At beginning of the running time, all weights ($\mu_{p}$) values are equal to zero for all data points. After the first iteration, the algorithm starts to update the k weights of each data point using the formula in equation~\ref{eq:2}. According to equation~\ref{eq:2},  $\delta_{i}$ is the previous weight of the cluster  $ i$ resulting from the previous frame, which will be updated using scale of weight $\Psi$ to update all memberships.\\
\begin{equation} \label{eq:2}
\delta_{i}=\delta_{i}+\Psi\; \; where \; \; 0<\Psi\leq 1.
\end{equation}\\
For each iteration, we update the weights of a data point by equations~\ref{eq:2} and~\ref{eq:3} where $\delta_{i}$  is weigh before normalization, and  $\tau$  is the number of iterations. The increasing rate of $\tau$ depends on frame rate that is addressed in section~\ref{sec:Results} and Figure~\ref{fig:7}. Therefore, iteration number is increasing around $fps_{t}$ per second.

\begin{equation}\label{eq:3}
\tau=\sum_{s=1}^{t}{fps_{t}}
\end{equation}

Equation~\ref{eq:4} presents a distance function of the clustering step, Euclidean distance between data points is used as similarity measure in many clustering algorithms including k-means. To consider the color difference of these points while clustering the data point of each frame, the RGB value is incorporated in the Euclidean distance between any two point's $v_{i}$ and $v_{c}$ as addressed in the following equation~\ref{eq:4}, where the scales $\alpha$ and $\delta$ insure that geometric distance and the color distance between two points are in the same order of magnitude. Experimentally, the best value of $\alpha$  is between 0.002 and 0.1. The scale of position is denoted by $\delta$ that is calculated from $\alpha$   using equation~\ref{eq:5}. By this equation, we define our new hybrid similarity measure as a function f of two terms; one in the Euclidean distance, $dist$, between two points $v_i$ and $v_c$ which are data point and centroid information, respectively.

\begin{align}\label{eq:4}
dist(v_{i},v_{c})=\sqrt{ 
\begin{pmatrix*}[l]
\delta^{2} 
\begin{pmatrix*}[l]
(X_{i}-X_{c})^{2}+\\
(Y_{i}-Y_{c})^{2}+\\
(Z_{i}-Z_{c})^{2} \\
\end{pmatrix*}+\\
\alpha ^{2} 
\begin{pmatrix*}[l]
(R_{i}-R_{c})^{2}+\\
(G_{i}-G_{c})^{2}+\\
(B_{i}-B_{c})^{2} \\
\end{pmatrix*}
\end{pmatrix*}
}\;\;\;\;,
\end{align}

\begin{equation}\label{eq:5}
\delta+\alpha\leq 1\; \; where\; \; 0<\alpha<1.
\end{equation}
After calculating the distance between a point and a cluster centroid using their  position and color as two different features, we need to incorporate the effect of normal vectors difference between them; thus, for the Euclidean distance, $dist$, between two points $v_i$ and $v_c$ which denoted each data point and centroid respectively. The angle between normal of the centroid and normal of each data point is denoted by  $\theta$.  $\gamma$ indicates the scale of normal for distance function. The best value of scale of normal,  $\gamma$, in our  experiment was found to be between 0.0001 and 0.01. The final similarity measure between the data points and the centroids that examine the similarity between their positions, colors and normal vectors is given by the following equation~\ref{eq:6}.
\begin{equation}\label{eq:6}
f(v_{i},v_{c})=dist(v_{i},v_{c})+\gamma(1-\cos(\theta(n_{i}-n_{c}))).
\end{equation}
\subsubsection{Update weight}\label{sec:update}
 the algorithm updates the weights according to equation~\ref{eq:2} and normalizes them for each data point according to equations~\ref{eq:7} and~\ref{eq:8}. Then, the algorithm assigns each data point the label of the cluster with the highest weight. With regard to equations~\ref{eq:7} and ~\ref{eq:8}, ($\mu_pi$) denotes the weight of cluster $i$  of data point $p$ and $\tau$ is the number of iterations which are calculated by the equation~\ref{eq:3}. The summation of k weights of each data point after normalization could be less than or equal to one.
\vfill

\begin{equation}\label{eq:7}
\mu_p=\frac{\sum_{i=0}^{\tau}{(\delta_{i})}}{\|\sum_{i=0}^{\tau}{(\delta_{i})}\|},
\end{equation}
 $where$
\begin{equation}\label{eq:8}
\sum_{i=0}^{k}{\mu_{p}^{i}}\leq 1.
\end{equation}

\begin{algorithm*}
\DontPrintSemicolon
  \caption{}
  \label{al:1}
 \While{ main\textcolor{blue} {\tcp{This is main loop that is starting at the binging and each iteration representing one frame}}}
 {
  \While{ all of data points\textcolor{blue} {\tcp{this loop represents the number of data points we have in each frame }}}
  {
  \eIf{Is it the first iteration \textcolor{blue} {\tcp{for the first iteration only we need to initialize the seeds}}}
  {
  Assign centroid by using K-means++ \textcolor{blue} {\tcp{ calculating the all centroid using K-means++}}
  Assign first weighted using regular labels\;\textcolor{blue} {\tcp{calculating the first label for each data point}}
   }{
       \While{ all of clusters\textcolor{blue} {\tcp{this loop start from 1 to k }}}{    
\begin{align}\label{eq:4}
dist(v_{i},v_{c})=\sqrt{ 
\begin{pmatrix*}[l]
\delta^{2} 
\begin{pmatrix*}[l]
(X_{i}-X_{c})^{2}+\\
(Y_{i}-Y_{c})^{2}+\\
(Z_{i}-Z_{c})^{2} \\
\end{pmatrix*}+\\
\alpha ^{2} 
\begin{pmatrix*}[l]
(R_{i}-R_{c})^{2}+\\
(G_{i}-G_{c})^{2}+\\
(B_{i}-B_{c})^{2} \\
\end{pmatrix*}
\end{pmatrix*}
}\nonumber
\end{align}
\textcolor{blue} {\tcc{calculating distance between each data point and centroid using position and color of each data point}}

\begin{equation}
f(v_{i},v_{c})=dist(v_{i},v_{c})+\gamma(1-\cos(\theta(n_{i}-n_{c})))\nonumber
\end{equation}
\textcolor{blue} {\tcc{incorporate the differences between angle of normal vectors of each data point and centroid}}

  \If{ $ f({v_i})  > f({V_{i+1}})$ \textcolor{blue} {\tcp{ condition of distance between previous frame and current frame }} }
  {
  
  \begin{equation}\label{eqn:deltai}
update\;\; labeled\;\; weights\;\; \delta_{i}=\delta_{i}+\Psi\; \; where \; \; 0<\Psi\leq 1 \nonumber
\end{equation}
\textcolor{blue} {\tcc{updating the weight of data point by scale of weights (one in our experiment)}}

\begin{equation}
\mu_p=\frac{\sum_{i=0}^{\tau}{(\delta_{i})}}{\|\sum_{i=0}^{\tau}{(\delta_{i})}\|}\nonumber
\end{equation}

\begin{equation}
\sum_{i=0}^{k}{\mu_{p}^{i}}\leq 1 \nonumber
\end{equation}
\textcolor{blue} {\tcc{normalize all weights for each data point where the summation of all weights should be less or equal than one }}

  }
  }
  }}
    \While{ all of data points \textcolor{blue} {\tcp{ assigning color to each data point for visualization  }}}
  {

\begin{equation}
C_{p}=\sum_{i=0}^{k}{(c_{i}\star \mu_{i})}\nonumber
\end{equation}
%%\textcolor{blue} {\tcc{assigning color to each data point by summation of unique cluster's color and weights of each data point }}
  }  
 }
\end{algorithm*}

%%%%%%%%%%%%%%%%%%%%%%%%%%%%%%%%%%%%%%%%%%%%%%%%%%%%%%%%%%%%%%%%%%%%%%%%%

%%%%%%%%%%%%%%%%%%%%%%%%%%%%%%%%%%%%%%%%%%%%%%%%%%%%%%%%%%%%%%%%%%%%%%
%%%%%%%%%%%%%%%%%%%%%%%%%%%%%%%%%%%%%%%%%%%%%%%%%%%%%%%%%%%%%%%%%%%%%%%%%

\subsection{Visualization} \label{sec:Visualization}
After calculating all k weights for all data points at each iteration, we illustrate the results as an object detection or segmentation by assigning each cluster a unique color $C_i$. Equation~\ref{eq:9} indicates assigning one of the k colors to each data point based on its weight $\mu$.  According to equation~\ref{eq:8}, the summation of all weights for data point is less than or equal to one, where $\mu$ is the weight of each cluster in a data point. That means, if we have a data point with k different weighted labels, the color of it is assigned by following equation:
\begin{equation}\label{eq:9}
C_{p}=\sum_{i=0}^{k}{(C_{i}\star \mu_{i})}
\end{equation}
As regards to Algorithm~\ref{al:1}, it has one main loop that contains the iteration of our system, the second loop contains clustering step that calculate and update weights. For each data points, the algorithm calculates its distance with respect to all centroids using RGB as color space, XYZ as points point position and $n_x$, $n_y$, $n_z$  as normal vector. After that algorithm updates the k weights of each data points after each iteration. Initially in the first iteration, k seeds are obtained by k-means++ and all weights labels are initialized equal to zero. For color assignment, all k weight of each point are used to give the point a label.

%%%%%%%%%%%%%%%%%%%%%%%%%%%%%%%%%%%%%%%%%%%%%%%%%%%%%%%%%%%%%%%%%%%%%%%%%
%%%%%%%%%%%%%%%%%%%%%%%%%%%%%%%%%%%%%%%%%%%%%%%%%%%%%%%%%%%%%%%%%%%%%%%%%
%%%%%%%%%%%%%%%%%%%%%%%%%%%%%%%%%%%%%%%%%%%%%%%%%%%%%%%%%%%%%%%%%%%%%%%%%
%%%%%%%%%%%%%%%%%%%%%%%%%%%%%%%%%%%%%%%%%%%%%%%%%%%%%%%%%%%%%%%%%%%%%%%%%

\section{Results}\label{sec:Results}
We test our algorithm using different number of clusters, k, and evaluate memory consumption and frame rate. Figure ~\ref{fig:7} indicates the frame rate experiments with  k between 2 and 100 along with the memory consumption of each experiment. The frame rate and memory consumption are tightly affected by the number of clusters. When we have large number of clusters, the algorithm needs more memory, and frame rate will be reduced. Figure~\ref{fig:3}  shows the result of weighted supervised learning with k = 7 clusters after several iterations; the memory consumption for seven clusters is 322 MB, and frame rate  is 7 $\pm  0.2$ frame per second. The figure~\ref{fig:4} shows the result of k=10 clusters where memory consumption is 344 MB, and frame rate is 6.5 $\pm   0.2$ frame per second. The algorithm implementation consumes multi-thread programming in Visual Studio 2015, to implement parallel processing for the prepossessing and clustering parts. All of the loops use the parallel computational  model  introduced by Microsoft API~\cite{RefWorks:99} assign including for each data point back-projection, normal calculation, mapping, and assigning color along with calculating the weights of each data point~\cite{RefWorks:103}. The used hardware running our algorithm plays another factor for evaluating the system.  The used CPU is capable of multi-thread programming and parallel processing, which is dual Xeon E5 family 2.29 GHz speed. It has 12 core and 24 logical processors. We do not use GPU in this application, but the GPU of the system is Nvidia Quadro K5000 with 4 GB GDDR5 speed~\cite{RefWorks:104}  with 32 GB memory and having a speed of 1333 MHz. In addition, we use Universal Serial Bus (USB) version 3.0 for  Kinect connection. Our input camera is Kinect V2 for Windows~\cite{RefWorks:95}.

%%%%%%%%%%%%%%%%%%%%%%%%%%%%%%%%%%%%%%%%%%%%%%%%%%%%%%%%%%%%%%%%%%%%%%%%%
%%%%%%%%%%%%%%%%%%%%%%%%%%%%%%%%%%%%%%%%%%%%%%%%%%%%%%%%%%%%%%%%%%%%%%%%%

\section{Discussion}\label{sec:Discussion}
With regard to our results, this algorithm provides a unique output that can be useful for researchers in computer graphics, computer vision, robotics, and other related fields. The weighted unsupervised learning for object detection proposed in this paper shows its capabilities to the smooth the output in noisy environments in real-time producing scene objects. The results shown in Figure~\ref {fig:7} indicate that this model can be real-time and has the capability for mobility applications with low memory consumption without the need for an expensive GPU. Using three features in clustering step, position, color and normal, gives us the capability to group similar data points to objects with accurate results. Our results show that the use of the weight to indicate the membership for each data point to a cluster and then for object detection is sufficient for detecting moving objects in noise environment, which was captured by an RGB-D camera.

%%%%%%%%%%%%%%%%%%%%%%%%%%%%%%%%%%%%%%%%%%%%%%%%%%%%%%%%%%%%%%%%%%%%%%%%%
%%%%%%%%%%%%%%%%%%%%%%%%%%%%%%%%%%%%%%%%%%%%%%%%%%%%%%%%%%%%%%%%%%%%%%%%%

\section{Conclusion and Future work}\label{sec:Conclusion}
In this paper, we proposed a weighted unsupervised learning for object detection framework using a single RGB-D camera. In the proposed method, we use CPU-based programming and multi-thread processing. The results are provided in real-time. In preprocessing step, we generate 3D data points, and after preprocessing, clustering step is applied by initialization of seeds and updating the weights of each data point. Finally assigning colors is used to illustrate object detection results. Another distinct contribution of this paper is segmentation of a particular object, particularly for moving objects that can be updated by each frame. The frame rate of this work is between 1 and 12 frame per second, and memory consumption of this algorithm is between 280 MB and 1 GB for different values of clusters. This algorithm can be applied in any object detection and segmentation application in the fields of computer graphics, robotics, vision, or for surveillance and object controlling. Future directions of this work can be summarized in extending this method to GPU based programming, increasing the performance and frame rate simultaneously.

\section*{Acknowledgment}

The authors want to acknowledge the help, resources and equipment that was provided by Professor James K. Hahn.



%\bibliographystyle{IEEEtran}

\balance
\section{appendix}\label{sec:appendix}
Converting the projective 3D position to real world with respect to camera is done by following equations~\ref{eq:10},~\ref{eq:11},~\ref{eq:12},~\ref{eq:13}, and~\ref{eq:14},  where field of view is denoted by $fov$, $W_x$, $W_y$ and $W_z$ is  world coordinate system of X, Y, and Z respectively, and $P_x$, $P_y$, and $P_z$ shows the projection parameters.

\newcommand{\scale}{\mathop{\mathrm{scale}}}

\begin{gather}
\scale_{x}=2\star \tan(\frac{fov_{x}}{2})\label{eq:10}\\
Scale_{y}=2\star \tan(\frac{fov_{y}}{2})\label{eq:11}\\
W_{x}=P_{z}\star Scale_{x}\star(\frac{P_{x}}{R_{w}}-0.5)\label{eq:12}\\
W_{y}=P_{z}\star Scale_{y}\star(\frac{P_{y}}{R_{w}}-0.5)\label{eq:13}\\
W_{z}=P_{z}\label{eq:14}
\end{gather}
$fov_{x}$ for Kinect 1is equal to 1.014468 and $fov_{y}$ is equal to 0.7898094 ~\cite{RefWorks:112}, and in many researches, people use these numbers that reduced the field of view, but as new resolution of Kinect V2 we use 1.22173047 for $fov_{x}$ and 1.0471975511 by changing degree to radian $Radian= \frac{Degree \star \pi}{180}$. The weight of each data point can be equal or bigger than zero, $\mu_{i}>0$ where $i\in [0,\; k-1]$ and k is the number of clusters.
The distance which is calculated in equation ~\ref{eq:4} and ~\ref{eq:6} and shown as $f(v_i,v_c )$, can be zero if and only if centroid and data point i are equal to each other, $i=c$. The scale of color and position as shown $\delta$ and $\alpha$, both are positive number and cannot be zero, and as we mention to that, summation of them is less than one. All results are calculated by sampling of one, so if we increase the sampling to more than one our resolution, and memory consumption will decrease and frame rate will be increased.
\end{document}